%% file: 0.main.tex
\pdfoutput=1
\documentclass[11pt]{article}

\usepackage[preprint]{acl}

\usepackage{times}
\usepackage{latexsym}

\usepackage[T1]{fontenc}
\usepackage[utf8]{inputenc}

\usepackage{microtype}

\usepackage{inconsolata}

\usepackage{graphicx}

\usepackage{booktabs}

\usepackage{amssymb}

\usepackage{tikz}
\usetikzlibrary{automata,positioning,shapes,calc}
\usetikzlibrary{arrows,calc,decorations.markings,math,arrows.meta}
\usetikzlibrary{fit,quotes,patterns,graphs,shapes.callouts}

\usepackage{pgfplots}
\pgfplotsset{scaled y ticks=false}

\usepackage{bm}
\input{math_commands}

\DeclareMathOperator*{\conv}{conv}
\newcommand\closedots{\makebox[1em][c]{.\hfil.\hfil.}}

\usepackage{algorithm}

\newcounter{xmpl}
\newenvironment{example}[1]
{%
\\
\noindent
\begin{minipage}{0.05\textwidth}\refstepcounter{xmpl}(\thexmpl\label{#1})\end{minipage}%
\begin{minipage}{0.95\textwidth}}
{\end{minipage}}

\usepackage{listings}
\newcommand{\cadec}{\textsc{Cadec}}
\newcommand{\sha}{\textsc{Share2013}}
\newcommand{\shb}{\textsc{Share2014}}

\title{A Fast and Sound Tagging Method\\for Discontinuous Named-Entity Recognition}

\author{Caio Corro  \\
  INSA Rennes, IRISA, Inria, CNRS, Université de Rennes\\
  \texttt{caio.corro@irisa.fr}
}

\date{}

\begin{document}
\maketitle
\begin{abstract}
We introduce a novel tagging scheme for discontinuous named entity recognition based on an explicit description of the inner structure of discontinuous mentions.
We rely on a weighted finite state automaton for both marginal and maximum \emph{a posteriori} inference.
As such, our method is sound in the sense that (1) well-formedness of predicted tag sequences is ensured via the automaton structure and (2) there is an unambiguous mapping between well-formed sequences of tags and (discontinuous) mentions.
We evaluate our approach on three English datasets in the biomedical domain,
and report comparable results to state-of-the-art while having a way simpler and faster model.
\end{abstract}

\input{1.introduction}
\input{2.tagging_scheme}
\input{3.algorithm}
\input{4.training}
\input{5.related_work}

\input{6.experiments}
\input{7.conclusion}

\section*{Acknowledgments}
I thank Vlad Niculae and François Yvon for their comments and suggestions.
I thank Lucas Ondel-Yang for the many discussions on finite state-automata that inspired the decoding algorithm described in this paper.
I thank Pierre Zweigenbaum for the help with Share datasets and the UMLS database.

Work partially done while I was a researcher at LISN and ISIR.
This work was granted access to the HPC/AI resources of IDRIS under the allocation 2024-AD011013727R1 made by GENCI.

\section*{Limitations}
The approach proposed in this paper cannot cover all form of discontinuities observed in the three datasets.
Indeed, some discontinuous mentions are composed of three parts or more.
However, they are rare so our results are still competitive.
Moreover, our contribution is focused on the general decoding approach that can be extended by future work.

Discontinuous NER datasets are scarce, therefore we are only able to experiment on three datasets in the biomedical domain in English.
We suspect this is due to a \emph{chicken or the egg} dilemma:
discontinuity are often not annotated as there are no easy plug-and-easy approach to predict them, and there is little NLP work in the domain as there are only a few datasets available for experiments.

During the evaluation of our approach, we observed that many mentions are missing in the gold annotation.
As such, all results reported on these datasets (including previous works) should be taken \emph{with a pinch of salt}.

\bibliography{custom}
%\bibliographystyle{acl_natbib}

%\onecolumn
\clearpage
\appendix

\input{10.nn}

\end{document}

%% file: math_commands.tex
\usepackage{amsmath,amsfonts,bm}

\def\eqref#1{equation~\ref{#1}}
\def\1{\bm{1}}

\def\vmu{{\bm{\mu}}}

\def\vw{{\bm{w}}}
\def\vx{{\bm{x}}}
\def\vy{{\bm{y}}}

\def\evw{{w}}

\DeclareMathAlphabet{\mathsfit}{\encodingdefault}{\sfdefault}{m}{sl}
\SetMathAlphabet{\mathsfit}{bold}{\encodingdefault}{\sfdefault}{bx}{n}

\newcommand{\R}{\mathbb{R}}

\DeclareMathOperator*{\argmax}{arg\,max}

%% file: 1.introduction.tex
\section{Introduction}

Named-entity recognition (NER) is a fundamental natural language processing (NLP) task that aims at identifying mentions of named entities in texts.
These mentions may for example refer to persons, organizations, locations or even dates, among others \cite{muc6,muc7ner}.
Over the years, this task has been extensively studied by the community, with contributions including decoding algorithms, neural network architectures, loss functions and methods for learning in different data availability situations, \emph{inter alia}.

There exists several variants of the NER problem,
among which the most studied are \emph{flat} NER and \emph{nested} NER.
The most common method for the flat case is \textsc{BIO} tagging \cite{ramshaw1995bio}, where each word in a sentence is tagged depending on whether it is the begining of a mention (\textsc{B}), inside a mention (\textsc{I}) or outside a mention (\textsc{O}).\footnote{See \citep{ratinov2009nerdesign} for other variants.}
This tagging scheme can be augmented to disambiguate types, \emph{e.g.}\ \textsc{BLoc} and \textsc{BPer}.
An important benefit of \textsc{BIO} tagging is that prediction has a linear time-complexity in the input length\footnote{It is quadratic in the number of tags, which depends on the number of possible mention types. However, types are not considered part of the input and are assumed to be fixed.} using the Viterbi algorithm \cite{forney1973viterbi}, contrary to concurrent approaches like semi-Markov models that have a quadratic time-complexity \cite{janssen2013semimarkov,ge2002semimarkov,sarawagi2004semimarkov}.

A less studied task is \emph{discontinuous} NER, where mentions are allowed to span discontinuous sequences of words.
This problem is especially important for biomedical NLP.
For example, pharmacovigilance aims to detect adverse drug reactions after a product is distributed in the market via automatic analysis of medical reports or social media \cite{berlin2008adr,coloma2013postmarketing}.
Mentions of adverse drug reactions naturally occur in non-contiguous sequences,
for example the sentence ``\texttt{The pain I was experiencing around the hipjoints was incredible}'' contains the mention ``\texttt{pain hipjoints}'' with a five word gap in the middle.

Several methods for discontinuous NER have been proposed in the literature, including transition models \cite{dai2020dner_transition} and other structured prediction approaches \cite{wang2021dner_clique,fei2021pointer,fei2022uniner}.
Unfortunately, they are more costly than \textsc{BIO} tagging and require specialized neural network architectures.
There have also been attempts to propose tagging schemes for discontinuous NER \cite{tang2013dner_tagging,tang2018recognizing,metke2016concept,muis2016graph},
but they all exhibit \emph{structural ambiguity} (see Section~\ref{sec:related_work}).

In this work,
we propose a novel tagging scheme for discontinuous NER that exploits the inner structure of discontinuous mentions.
Contrary to previous attempts, our approach is \emph{sound} in the sense that:
(1)~there is no encoding ambiguity between sets of mentions and sequences of tags (\emph{i.e.}\ there is a one-to-one mapping between the two representations); and (2)~our prediction algorithm is constrained to predict only well-formed sequences of tags (\emph{i.e.}\ we can always reconstruct a set of mentions from a predicted tag sequence).
To ensure well-formedness of predictions, we propose an algorithm based on inference in a weighted finite-state automaton.
Using our approach, the time complexity of maximum \emph{a posteriori} inference for prediction is linear in the length of the input.
Moreover, our algorithm can be very efficiently implemented on GPU for batched inference \cite{argueta2017gpu,rush2020torchstruct}.

Our contributions can be summarized as follows:
\begin{itemize}
    \item We propose to decompose discontinuous mentions in a new two-layer representation;
    \item We propose a novel tagging scheme for this representation together with a linear-time tagging algorithm that ensures well-formedness of predictions;
    \item We explain how labels in the inner structures can be inferred during training when the information is not available in the data;
    \item We experiment on three English datasets and report competitive results while having a much faster model.
\end{itemize}
Our implementation is publicly available.\footnote{\url{https://github.com/FilippoC/disc-ner-tagging}}
Importantly, our decoding algorithm and all our loss functions can be used as a drop-in replacements in any \textsc{BIO} tagger.
As such, any future research in the \textsc{BIO} tagging field may also be evaluated on discontinuous NER at no extra cost.

%% file: 2.tagging_scheme.tex
\section{Reduction to Word Tagging}

In this section, we explain how we map discontinuous mentions into a two-layer representation that allows us to derive a new tagging scheme.
Although this transformation is generic, for ease of exposition we illustrate it on the particular case of adverse drug reactions.

\subsection{Inner Structure of Mentions}

Discontinuous mentions of adverse drug reactions (ADR) and disorders in biomedical NER mainly result from two linguistic phenomena.
Firstly, mentions may be expressed as the combination of two non-contiguous syntactic constituents, due to linguistic word order rules.
In the following example of an ADR, the discontinuity is caused by the verb position constraint in English:
\begin{example}{ex:toes}
    \input{examples/subj-obj}
\end{example}
Secondly, many languages allow alternative sentential structures for coordinations, including construction based on deletion operations.
For example, consider the two following sentences:
\begin{example}{ex:reduction1}
    \input{examples/reduction1}
\end{example}
\begin{example}{ex:reduction2}
    \input{examples/reduction2}
\end{example}
The repeated element is eliminated in the second one, leading to the presence of a discontinuous mention, a phenomenon called coordination reduction \cite{lakoff1969reduction}.
Although the underlying linguistic structures are different, we will treat both cases in the same way.

\textbf{Change of representation.}
In practice, discontinuous mentions exhibit an inner structure.
For example, a discontinuous ADR can be decomposed into a \emph{body part} and an \emph{event}.
As such, we propose to transform discontinuous mentions into a two-layer representation:
\begin{itemize}
    \item Upper layers identify \emph{sets of mentions}; 
    \item Lower layers identify \emph{typed components}.
\end{itemize}
We restrict the number of types for components to be equal to two.
The previous example is converted as follows:
\begin{example}{ex:inner}
    \input{examples/inner}
\end{example}
Note that the two mentions do not explicitly appear in this new representation.
Nevertheless, the opposite transformation is trivial:
to rebuild all discontinuous mention in a discontinuous set,
we simply take the Cartesian product between the two sets of typed components,
\emph{e.g.}
$$\small
\underbrace{
    \vphantom{ % for underbrace vertical alignment
        \begin{array}{@{}l@{}}
             \texttt{arms},  \\
             \texttt{shoulders} 
        \end{array}
    }
    \{\texttt{pain in}\}
}_{\substack{\text{Components} \\ \text{typed \textsc{Event}}}}
\times
\underbrace{
\left\{
\begin{array}{@{}l@{}}
     \texttt{arms},  \\
     \texttt{shoulders} 
\end{array}
\right\}
}_{\substack{\text{Components} \\ \text{typed \textsc{Part}}}}
\mapsto
\underbrace{
    \left\{
    \begin{array}{@{}l@{}}
         \texttt{pain in arms},  \\
         \texttt{pain in shoulders} 
    \end{array}
    \right\}
}_{\substack{\text{Reconstructed}\\\text{discontinuous mentions}}}.
$$
Note that this can result in some of the mentions being continuous, as in Example~(\ref{ex:inner}).

One obvious issue is that component types are not annotated in datasets.
We consider two solutions to tackle this challenge.
First, we can use unsupervised and weakly-supervised learning methods to infer component types during training, as explained in Section~\ref{sec:weaklysup}.
Second, we can use component types to mark if they share the same type as the leftmost one, no matter whether they refer to a body part of an event.
In this setting, Examples~(\ref{ex:toes}) and~(\ref{ex:reduction2}) are annotated as follows:
\begin{example}{ex:leftmost2}
    \input{examples/leftmost2}
\end{example}
\begin{example}{ex:leftmost1}
    \input{examples/leftmost1}
\end{example}
In other words, component types do not convey semantic information, only structural information.

\textbf{Continuous mentions.}
There exists two forms of continuous mentions.
First, continuous mentions that share one or more words with at least one other mention.
In this case, we split the mention and we process it as described above.
Second, there are continuous mentions that do no share any word with other mentions, see Example~(\ref{ex:reduction1}).
In principle, we could also transform these mentions in the two layers representation.
However, not only we lack information about component types but we do not even know where to split them!
In Example~(\ref{ex:reduction2}), we know that ``\texttt{pain in arms}'' should be splitted into ``\texttt{pain in}'' and ``\texttt{arms}'' as the first two words are shared with another mention.
But for the two continuous mentions in Example~(\ref{ex:reduction1}), we do not have such information.
Therefore, in this case, we treat them as standard continuous ones.

\textbf{Nested NER.}
Although \citet{dai2020dner_transition} suggested the use of nested NER models for discontinuous NER
using a similar yet different representation,
we argue that the two problems are different:
\begin{itemize}
    \item The structures that we consider are not recursive, contrary to nested mentions, \emph{e.g.}\ ``\texttt{[The president of [the United States of [America]]]}'';
    \item The components are highly constrained, \emph{e.g.}\ a set of ADRs must contain at least one body part and one event;
    \item The span of a set of mentions is fixed by its components: it begins (resp.\ ends) at the same word as its leftmost (resp.\ rightmost) component.
\end{itemize}
Therefore, we instead propose a tagging scheme tailored to discontinuous NER.

\textbf{Beyond the biomedical domain.}
Our approach can be applied to other domains,
\emph{e.g.}\ we can transform the following mentions into our representation by differentiating first and last names:
\begin{example}{ex:name}
    \input{examples/name}
\end{example}
Unfortunately, these discontinuities have not been annotated in standard datasets.\footnote{\citet{wang2023corec} automatically extracted coordination structures from syntactic structures. However, note that (1)~the resulting dataset does not contains discontinuous mentions that we are interested in and (2)~conjunction reduction cannot always be inferred from the syntactic structure \cite{lakoff1969reduction,lechner2000reductioin,wilder2018reduction}.}

\subsection{Tagging Scheme}
\label{sec:scheme}

\input{figures/tag_sequence}

We now explain how we transform the two-layer structure into a sequence of tags.
Without loss of generality, we assume that mentions are untyped in the original corpus, as practical datasets for discontinuous NER contain a single mention type.\footnote{It is trivial to augment the set of tags with types if necessary, as done for standard $\textsc{BIO}$ tagging.}
Moreover, we define the component types as $\textsc{x}$ and $\textsc{y}$ (\emph{e.g.}\ \emph{body part} and \emph{event} in previous examples) to simplify notation and treat in a similar way semantic and structural component types.

Our approach requires 10 tags.
First, the 3 tags $\textsc{CB}$, $\textsc{CI}$ and $\textsc{O}$ are used in a similar way to BIO tags.
$\textsc{CB}$ and $\textsc{CI}$ are used to identify first and following words in a continuous mention, respectively.
The tag $\textsc{O}$ is used to mark words that are neither part of a continuous mention or in the span of a set of mentions. In Example~(\ref{ex:reduction1}), word ``\texttt{and}'' is tagged with $\textsc{O}$ whereas in Example~(\ref{ex:reduction2}) it \emph{is not} tagged with $\textsc{O}$.
This is due to the fact that in the second example, after transformation into the two layers representation, the word ``\texttt{and}'' will appear inside a set of mentions, see Example~(\ref{ex:inner}).

Second, tags to identify set of mentions and their components are of the form \textsc{*-*} where:
\begin{itemize}
    \item the left-hand side is used to identify the span of the set of mentions, and can therefore take values \textsc{DB} (first word of the span) and \textsc{DI} (other words of the span);
    \item the right-hand side is used to identify typed components, and can take values \textsc{Bx}, \textsc{Ix}, \textsc{By}, \textsc{Iy} and \textsc{O}.
\end{itemize}
The 7 tags used for discontinuous mentions are 
\textsc{DB-Bx}, \textsc{DB-By},
\textsc{IB-Bx}, \textsc{IB-By},
\textsc{IB-Ix}, \textsc{IB-Iy}
and \textsc{IB-O}.
Note that the leftmost word in a set of mentions must also be the beginning of a component,
so the following combinations \emph{are not} part of the tagset: \textsc{DB-Ix}, \textsc{DB-Iy} and \textsc{DB-O}.
Figure~\ref{fig:tag_sequence} shows an example of tag conversion.

Importantly, any sequence of tags is \emph{well-formed} if and only if:
\begin{enumerate}
    \item All $\textsc{CI}$ tags are preceded by either $\textsc{BI}$ or $\textsc{CI}$, as standard in $\textsc{BIO}$ tagging;
    \item All $\textsc{DI-*}$ tags must be preceded by either $\textsc{DB-*}$ or $\textsc{DI-*}$;
    \item All $\textsc{*-Ix}$ tags must be preceded by either $\textsc{*-Bx}$ or $\textsc{*-Ix}$ (and similarly for the $\textsc{y}$ type);
    \item A set of mentions must contain at least one component typed \textsc{x} and one typed \textsc{y}, that is it must contain at least one word tagged with \textsc{*-Bx} and one with \textsc{*-By}.
    \item A set of mentions must not yield a single continuous mention after reconstruction, \emph{i.e.}\ the following sequence of tags is forbidden:
    \begin{example}{ex:forbidden}
        \input{examples/forbidden}
    \end{example}
    as it would introduce ambiguity in the encoding of continuous mentions;
    \item A discontinuous mention cannot end with tag $\textsc{DI-O}$, as this would results in the span of a set of mentions that do not end with the same word as its rightmost component.\footnote{The analogous constraint on the first word is implicitly enforced by the absence of a $\textsc{DB-O}$ tag in the tagging scheme.}
\end{enumerate}

%% file: examples/subj-obj.tex
\begin{tikzpicture}[
    every node/.style={
        rectangle,
        inner xsep=0cm,
        inner ysep=0.1cm,
        text height=1.5ex,
        text depth=.25ex,
    }
]
    \node (toes) [rectangle] {\texttt{toes}};
    \node (are) [rectangle, right=0.2cm of toes] {\texttt{are}};
    \node (painful) [rectangle, right=0.2cm of are] {\texttt{painful}};

    \draw (toes.north west) --coordinate (part1)  (toes.north east);
    \draw (painful.north west) --coordinate (part2) (painful.north east);

    \coordinate[above=0.2cm of part1] (part1_b);
    \draw[dashed] (part1) -- (part1_b);
    \coordinate[above=0.2cm of part2] (part2_b);
    \draw[dashed] (part2) -- (part2_b);

    \draw[dashed] (part1_b) --node[yshift=+0.25cm] {\textsc{Adr}} (part2_b);

\end{tikzpicture}

%% file: examples/reduction1.tex
\begin{tikzpicture}[
    every node/.style={
        rectangle,
        inner xsep=0cm,
        inner ysep=0.1cm,
        text height=1.5ex,
        text depth=.25ex,
    }
]
    % weakness in arms and weakness in shoulders
    \node (pain) [rectangle] {\texttt{pain}};
    \node (in) [rectangle, right=0.2cm of pain] {\texttt{in}};
    \node (arms) [rectangle, right=0.2cm of in] {\texttt{arms}};
    \node (and) [rectangle, right=0.2cm of arms] {\texttt{and}};
    \node (pain2) [rectangle, right=0.2cm of and] {\texttt{pain}};
    \node (in2) [rectangle, right=0.2cm of pain2] {\texttt{in}};
    \node (shoulders) [rectangle, right=0.2cm of in2] {\texttt{shoulders}};

    \draw(pain.north west)--node[yshift=0.25cm]{\textsc{Adr}} (arms.north east);
    \draw(pain2.north west)--node[yshift=0.25cm]{\textsc{Adr}} (shoulders.north east);
\end{tikzpicture}

%% file: examples/reduction2.tex
\begin{tikzpicture}[
    every node/.style={
        rectangle,
        inner xsep=0cm,
        inner ysep=0.1cm,
        text height=1.5ex,
        text depth=.25ex,
    }
]
    \node (r_pain) [rectangle] {\texttt{pain}};
    \node (r_in) [rectangle, right=0.2cm of r_pain] {\texttt{in}};
    \node (r_arms) [rectangle, right=0.2cm of r_in] {\texttt{arms}};
    \node (r_and) [rectangle, right=0.2cm of r_arms] {\texttt{and}};
    \node (r_shoulders) [rectangle, right=0.2cm of r_and] {\texttt{shoulders}};

    \draw(r_pain.north west)--node[yshift=0.25cm]{\textsc{Adr}} (r_arms.north east);

    \draw (r_pain.south west) --coordinate (part1)  (r_in.south east);
    \draw (r_shoulders.south west) --coordinate (part2)  (r_shoulders.south east);

    \coordinate[below=0.2cm of part1] (part1_b);
    \draw[dashed] (part1) -- (part1_b);
    \coordinate[below=0.2cm of part2] (part2_b);
    \draw[dashed] (part2) -- (part2_b);

    \draw[dashed] (part1_b) --node[yshift=-0.25cm] {\textsc{Adr}} (part2_b);
\end{tikzpicture}

%% file: examples/inner.tex
\begin{tikzpicture}[
    every node/.style={
        rectangle,
        inner xsep=0cm,
        inner ysep=0.1cm,
        text height=1.5ex,
        text depth=.25ex,
    }
]
    \node (r_pain) [rectangle] {\texttt{pain}};
    \node (r_in) [rectangle, right=0.2cm of r_pain] {\texttt{in}};
    \node (r_arms) [rectangle, right=0.2cm of r_in] {\texttt{arms}};
    \node (r_and) [rectangle, right=0.2cm of r_arms] {\texttt{and}};
    \node (r_shoulders) [rectangle, right=0.2cm of r_and] {\texttt{shoulders}};

    \draw(r_pain.north west)--node[yshift=0.25cm]{\textsc{Event}} (r_in.north east);
    \draw(r_arms.north west)--node[yshift=0.25cm]{\textsc{Part}} (r_arms.north east);
    \draw(r_shoulders.north west)--node[yshift=0.25cm]{\textsc{Part}} (r_shoulders.north east);
    \draw($(r_pain.north west)+(0,15pt)$)--node[yshift=0.25cm]{\textsc{Adr(s)}} ($(r_shoulders.north east)+(0,15pt)$);
\end{tikzpicture}

%% file: examples/leftmost2.tex
\begin{tikzpicture}[
    every node/.style={
        rectangle,
        inner xsep=0cm,
        inner ysep=0.1cm,
        text height=1.5ex,
        text depth=.25ex,
    }
]
    \node (r_pain) [rectangle] {\texttt{toes}};
    \node (r_in) [rectangle, right=0.2cm of r_pain] {\texttt{are}};
    \node (r_arms) [rectangle, right=0.2cm of r_in] {\texttt{painful}};
    
    \draw(r_pain.north west)--node[yshift=0.25cm]{\textsc{First}} (r_pain.north east);
    \draw(r_arms.north west)--node[yshift=0.25cm]{\textsc{Other}} (r_arms.north east);
    \draw($(r_pain.north west)+(0,15pt)$)--node[yshift=0.25cm]{\textsc{Adr(s)}} ($(r_arms.north east)+(0,15pt)$);
\end{tikzpicture}

%% file: examples/leftmost1.tex
\begin{tikzpicture}[
    every node/.style={
        rectangle,
        inner xsep=0cm,
        inner ysep=0.1cm,
        text height=1.5ex,
        text depth=.25ex,
    }
]
    \node (r_pain) [rectangle] {\texttt{pain}};
    \node (r_in) [rectangle, right=0.2cm of r_pain] {\texttt{in}};
    \node (r_arms) [rectangle, right=0.2cm of r_in] {\texttt{arms}};
    \node (r_and) [rectangle, right=0.2cm of r_arms] {\texttt{and}};
    \node (r_shoulders) [rectangle, right=0.2cm of r_and] {\texttt{shoulders}};

    \draw(r_pain.north west)--node[yshift=0.25cm]{\textsc{First}} (r_in.north east);
    \draw(r_arms.north west)--node[yshift=0.25cm]{\textsc{Other}} (r_arms.north east);
    \draw(r_shoulders.north west)--node[yshift=0.25cm]{\textsc{Other}} (r_shoulders.north east);
    \draw($(r_pain.north west)+(0,15pt)$)--node[yshift=0.25cm]{\textsc{Adr(s)}} ($(r_shoulders.north east)+(0,15pt)$);
\end{tikzpicture}

%% file: examples/name.tex
\begin{tikzpicture}[
    every node/.style={
        rectangle,
        inner xsep=0cm,
        inner ysep=0.1cm,
        text height=1.5ex,
        text depth=.25ex,
    }
]
    \node (Meg) [rectangle] {\texttt{Meg}};
    \node (and) [rectangle, right=0.2cm of Meg] {\texttt{and}};
    \node (Jack) [rectangle, right=0.2cm of and] {\texttt{Jack}};
    \node (White) [rectangle, right=0.2cm of Jack] {\texttt{White}};

    \draw(Jack.south west)--node[yshift=-0.25cm]{\textsc{Per}} (White.south east);

    \draw (Meg.north west) --coordinate (part1)  (Meg.north east);
    \draw (White.north west) --coordinate (part2) (White.north east);

    \coordinate[above=0.2cm of part1] (part1_b);
    \draw[dashed] (part1) -- (part1_b);
    \coordinate[above=0.2cm of part2] (part2_b);
    \draw[dashed] (part2) -- (part2_b);

    \draw[dashed] (part1_b) --node[yshift=+0.25cm] {\textsc{Per}} (part2_b);

\end{tikzpicture}

%% file: figures/tag_sequence.tex
\begin{figure*}[t]
\centering

\begin{tikzpicture}[
    every node/.style={
        rectangle,
        inner xsep=0cm,
        inner ysep=0.1cm,
        text height=1.5ex,
        text depth=.25ex,
    }
]

% Chronic fatigue together with swollen and stiff knees and elbows
\node (chronic) [rectangle] {\texttt{Chronic}};
\node (fatigue) [rectangle, right=0.4cm of chronic] {\texttt{fatigue}};
\node (together) [rectangle, right=0.4cm of fatigue] {\texttt{together}};
\node (with) [rectangle, right=0.4cm of together] {\texttt{with}};
\node (swollen) [rectangle, right=0.4cm of with] {\texttt{swollen}};
\node (and1) [rectangle, right=0.4cm of swollen] {\texttt{and}};
\node (stiff) [rectangle, right=0.4cm of and1] {\texttt{stiff}};
\node (knees) [rectangle, right=0.4cm of stiff] {\texttt{knees}};
\node (and2) [rectangle, right=0.4cm of knees] {\texttt{and}};
\node (left) [rectangle, right=0.4cm of and2] {\texttt{left}};
\node (elbows) [rectangle, right=0.4cm of left] {\texttt{elbows}};
\node [rectangle, right=0cm of elbows] {\texttt{.}};

\draw(chronic.north west)--node[yshift=0.25cm]{\textsc{Adr}} (fatigue.north east);

\draw(stiff.north west)--node[yshift=0.25cm]{\textsc{Adr}} (knees.north east);

\draw ($(stiff.north west)+(0,0.6)$) --coordinate (part1)  ($(stiff.north east)+(0,0.6)$);
\draw ($(left.north west)+(0,0.6)$) --coordinate (part2) ($(elbows.north east)+(0,0.6)$);
\coordinate[above=0.2cm of part1] (part1_b);
\draw[dashed] (part1) -- (part1_b);
\coordinate[above=0.2cm of part2] (part2_b);
\draw[dashed] (part2) -- (part2_b);
\draw[dashed] (part1_b) --node[yshift=+0.25cm] {\textsc{Adr}} (part2_b);

\draw ($(swollen.south west)-(0,0.8)$) --coordinate (part1)  ($(swollen.south east)-(0,0.8)$);
\draw ($(left.south west)-(0,0.8)$) --coordinate (part2)  ($(elbows.south east)-(0,0.8)$);
\coordinate[below=0.2cm of part1] (part1_b);
\draw[dashed] (part1) -- (part1_b);
\coordinate[below=0.2cm of part2] (part2_b);
\draw[dashed] (part2) -- (part2_b);
\draw[dashed] (part1_b) --node[yshift=-0.25cm] {\textsc{Adr}} (part2_b);

\draw (swollen.south west) --coordinate (part1)  (swollen.south east);
\draw (knees.south west) --coordinate (part2)  (knees.south east);
\coordinate[below=0.2cm of part1] (part1_b);
\draw[dashed] (part1) -- (part1_b);
\coordinate[below=0.2cm of part2] (part2_b);
\draw[dashed] (part2) -- (part2_b);
\draw[dashed] (part1_b) --node[yshift=-0.25cm] {\textsc{Adr}} (part2_b);

\node (r_chronic) [rectangle, below=3cm of chronic] {\texttt{Chronic}};
\node (r_fatigue) [rectangle, right=0.4cm of r_chronic] {\texttt{fatigue}};
\node (r_together) [rectangle, right=0.4cm of r_fatigue] {\texttt{together}};
\node (r_with) [rectangle, right=0.4cm of r_together] {\texttt{with}};
\node (r_swollen) [rectangle, right=0.4cm of r_with] {\texttt{swollen}};
\node (r_and1) [rectangle, right=0.4cm of r_swollen] {\texttt{and}};
\node (r_stiff) [rectangle, right=0.4cm of r_and1] {\texttt{stiff}};
\node (r_knees) [rectangle, right=0.4cm of r_stiff] {\texttt{knees}};
\node (r_and2) [rectangle, right=0.4cm of r_knees] {\texttt{and}};
\node (r_left) [rectangle, right=0.4cm of r_and2] {\texttt{left}};
\node (r_elbows) [rectangle, right=0.4cm of r_left] {\texttt{elbows}};
\node [rectangle, right=0cm of r_elbows] {\texttt{.}};

\draw(r_chronic.north west)--node[yshift=0.25cm]{\textsc{Adr}} (r_fatigue.north east);

\draw(r_swollen.north west)--node[yshift=0.25cm]{\textsc{Event}} (r_swollen.north east);
\draw(r_stiff.north west)--node[yshift=0.25cm]{\textsc{Event}} (r_stiff.north east);
\draw(r_knees.north west)--node[yshift=0.25cm]{\textsc{Part}} (r_knees.north east);
\draw(r_left.north west)--node[yshift=0.25cm]{\textsc{Part}} (r_elbows.north east);

\draw($(r_swollen.north west)+(0,15pt)$)--node[yshift=0.25cm]{\textsc{Adr(s)}} ($(r_elbows.north east)+(0,15pt)$);

\node (t_chronic) [rectangle, below=0cm of r_chronic] {\textsc{CB}};
\node (t_fatigue) [rectangle, below=0cm of r_fatigue] {\textsc{CI}};
\node (t_together) [rectangle, below=0cm of r_together] {\textsc{O}};
\node (t_with) [rectangle, below=0cm of r_with] {\textsc{O}};
\node (t_swollen) [rectangle, below=0cm of r_swollen] {\textsc{DB-Bx}};
\node (t_and1) [rectangle, below=0cm of r_and1] {\textsc{DI-O}};
\node (t_stiff) [rectangle, below=0cm of r_stiff] {\textsc{DI-Bx}};
\node (t_knees) [rectangle, below=0cm of r_knees] {\textsc{DI-By}};
\node (t_and2) [rectangle, below=0cm of r_and2] {\textsc{DI-O}};
\node (t_left) [rectangle, below=0cm of r_left] {\textsc{DI-By}};
\node (t_elbows) [rectangle, below=0cm of r_elbows] {\textsc{DI-Iy}};

\end{tikzpicture}

\caption{%
\textbf{(Top)}
Sentence with its original annotation.
It contains two continuous mentions (``\texttt{Chronic fatigue}'' and ``\texttt{stiff knees}'') and
three discontinuous mentions (``\texttt{swollen knees}'', ``\texttt{swollen left elbows}'' and ``\texttt{stiff left elbows}'').
\textbf{(Bottom)}
Sentence annotated with our two-layer representation and the associated tag sequence.
}
\label{fig:tag_sequence}

\end{figure*}
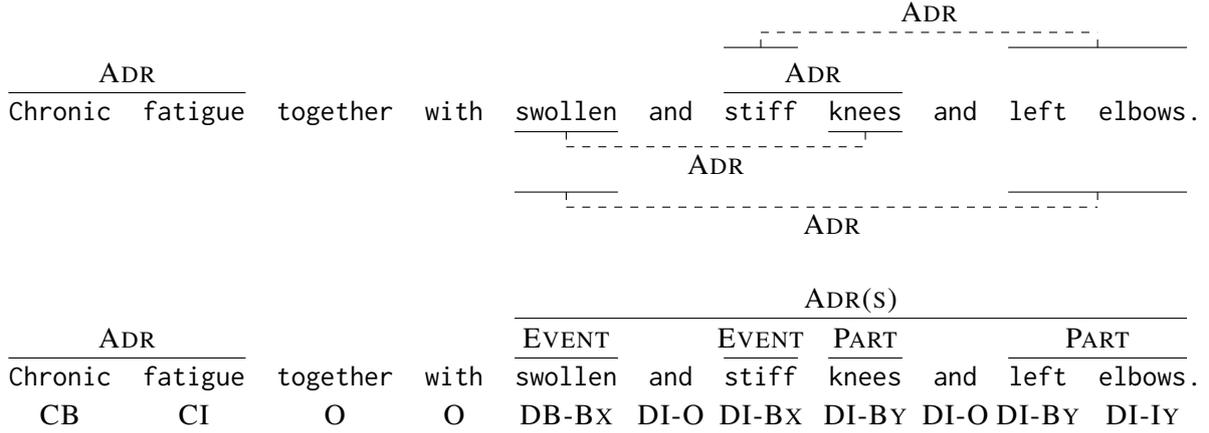

%% file: examples/forbidden.tex
\begin{tikzpicture}[
    every node/.style={
        rectangle,
        inner xsep=0cm,
        inner ysep=0.1cm,
        text height=1.5ex,
        text depth=.25ex,
    }
]
    % weakness in arms and weakness in shoulders
    \node (some) [rectangle] {\texttt{some}};
    \node (pain) [rectangle, right=0.7cm of some] {\texttt{pain}};
    \node (in) [rectangle, right=0.7cm of pain] {\texttt{in}};
    \node (arms) [rectangle, right=0.7cm of in] {\texttt{arms}};
    \node (and) [rectangle, right=0.7cm of arms] {\texttt{and}};

    \node (t_some) [rectangle, below=0cm of some] {\textsc{O}};
    \node (t_pain) [rectangle, below=0cm of pain] {\textsc{DB-Bx}};
    \node (t_in) [rectangle, below=0cm of in] {\textsc{DI-Ix}};
    \node (t_arms) [rectangle, below=0cm of arms] {\textsc{DI-Iy}};
    \node (t_and) [rectangle, below=0cm of and] {\textsc{O}};

    \draw[dashed] ($(some.north west)+(0, 0.2em)$) -- ($(and.south east |- t_and.south east)+(0, -0.2em)$);
    \draw[dashed] ($(and.north east)+(0, 0.2em)$) -- ($(some.south west |- t_some.south west)+(0, -0.2em)$);
\end{tikzpicture}

%% file: 3.algorithm.tex
\section{Decoding Algorithm}
\label{sec:decoding}

Without loss of generality, we assume all sentences have $n$ words.
Let $T$ be the tagset, $X$ be the set of sentences and $Y$ the set of well-formed tag sequences.
We represent a sequence of tags $\vy \in Y$ as a binary vector with $n  |T|$ entries,
where each entry is associated with a tag and a word, \emph{i.e.}\ $\vy \in \{0, 1\}^{n|T|}$.
If the value of an entry is 1 (resp.\ 0),
the associated tag is assigned to the associated word (resp.\ not assigned).
Note that $Y \subset \{0, 1\}^{n|T|}$ is a strict subset of all such vectors, as each word must be assigned exactly one tag and that the resulting tag sequence must satisfy the constraints described in Section~\ref{sec:scheme}.

Let $f_\theta: X \to \R^{n|T|}$ be a neural network parameterized by $\theta$.
We define the probability of a tag sequence $\vy \in Y$ given the input $\vx$ as a Boltzmann-Gibbs distribution (or \emph{softmax} over structures):
\begin{align}
    p_\theta(\vy | \vx) &= \exp\big(~\langle\vy, f_\theta(\vx) \rangle - A_Y(f_\theta(\vx))~\big)\,, \nonumber
\intertext{%
where $\langle \cdot, \cdot\rangle$ denotes the dot product and $A_Y$ is the log-partition function ensuring that the distribution is correctly normalized:}
    \label{eq:lognorm}
    A_Y(\vw) &= \log \sum_{\vy \in Y} \exp~\langle \vy, \vw\rangle\,.\\
\intertext{%
Computing $A_Y(\vw)$ is called \emph{marginal inference} due to its link with marginal probabilities \cite{wainwright2008exponential}.
Computing the most probable output is reduced to computing:
}
    \label{eq:map}
    \widehat\vy_\theta(\vx) &= \argmax_{\vy \in Y}~\langle \vy, f_\theta(\vx)\rangle\,,
\end{align}
called \emph{maximum a posteriori (MAP) inference}.

In practice, we need to compute the term in Equation~(\ref{eq:lognorm}) for training the model and the term in Equation~(\ref{eq:map}) for prediction.
The difficulty stems from the restriction (in the sum and in the $\argmax$ search space) to the set of well-formed outputs $Y$.
We follow a long tradition in NLP \cite[\emph{inter alia}]{koskenniemi1990fstparsing,mohri1996fsa,karttunen1996re,kanthak2004fsa,tromble2006globalfst,rastogi2016neuralfst,lin2019neuralfst,papay2022constraining} and rely on a finite-state automaton to solve these inference problems.

\subsection{Finite-State Automata}

\textbf{Definitions.}
Weighted Finite State Automata (WFSA) are generalization of FSA \cite{eilenberg1974automata} that include weights on their transitions.
Formally, a WFSA over $\R$ is a 5-tuple $( \Sigma, Q, E, i, F )$ where:
\begin{itemize}
    \item $\Sigma$ is a finite alphabet with $\epsilon \notin \Sigma$;
    \item $Q$ is the set of states;
    \item $E \subseteq Q \times \Sigma^* \times \R \times Q$ is the set of weighted transitions, where $(q, \sigma, w, r) \in E$ is a transition from state $q$ to state $r$ emitting symbol(s) $\sigma$ with weight $w$;
    \item $i \in Q$ is an initial state and $F \subseteq Q$ are final states.
\end{itemize}
Symbol $\epsilon$ is used for transitions that emit nothing.
A WFSA is $\epsilon$-free if there is no $\epsilon$-transition.
A \emph{valid path} is a path starting at $i$ and ending at any state in $F$.
A path emits a sequence of symbols, and has a weight equal to the sum of the transition weights it contains.
The language of a WFSA is the set of emissions along all possible valid paths.

\textbf{Algorithms.}
Given an acyclic WFSA,
the path of maximum weight, Equation~(\ref{eq:map}), and the log-sum-exp of all valid paths, Equation~(\ref{eq:lognorm}), can be computed using variants of the Viterbi algorithm \cite{forney1973viterbi} and the Forward algorithm \cite{baum1972forward}, respectively.
These algorithms are in fact identical, but defined over different semirings \cite{goodman1999semiring}:
the tropical semiring for the Viterbi and the thermodynamic semiring \cite{marcolli2014thermodynamic} for the Forward.
We refer to \cite[Section~3]{mohri2009alg} for an in-depth introduction.
The time complexity of both algorithms is $\mathcal O(|E|)$ if a topological ordering of states is known.

\textbf{Application to sequence tagging.}
We follow previous work and use the intersection of two WFSAs to constraint tag sequences \cite{koskenniemi1990fstparsing,koskenniemi1992compiling}.
The \emph{grammar automaton} $\mathcal G \triangleq (T, Q, E, i, F)$ is a cyclic WFSA whose language is the set of all well-formed tag sequences (of any length).
We assume $G$ is $\epsilon$-free and deterministic.\footnote{Procedures to determinize and remove $\epsilon$-transitions can be found in \citet[Section 2.3.5 and 2.5.5]{hopcroft2001automata}.}
Without loss of generality, we fix all transition weights to 0.
The \emph{sentence automaton} $\mathcal S \triangleq (T, Q', E', i', F')$ is an acyclic FSA that represents all possible (not necessarily valid) analyses for a given sentence of $n$ words.
States are $Q' \triangleq \{0, ..., n\}$ and transitions are:
$$
E' \triangleq \big\{
    (i-1, t, \evw_{(i, t)}, i) ~|~ i \in \{1 \closedots n\} \land t \in T
\big\}
$$
where $\evw_{(i, t)}$ is the weight associated with tagging word at position $i$ with tag $t$.
Initial and final states are $i' \triangleq 0$ and $F' \triangleq \{n\}$.
This WFSA contains $n|T|$ transitions, and each transition correspond to tagging a given word with a given tag.
By construction, it is always deterministic and $\epsilon$-free.

We denote $G \cap S$ the intersection of $G$ and $S$ \citep[Section 4.2.1]{hopcroft2001automata} composed of states $Q'' \triangleq Q \times Q'$, transitions
$$
E''\!\triangleq\!\left\{
    ((i\!-\!1,p), t, \evw_{(i, t)}, (i, q))
    \middle|
    \begin{array}{@{}l@{}}
    i \in \{1 \closedots n\} \land \\
    (p, t, 0, q)\!\in\!E
    \end{array}
\right\}\!,
$$
initial state $i'' \triangleq (i, i')$ and final states $F'' \triangleq F \times F'$.
Then, all valid paths in $G \cap S$ are well-formed sequences of tags for the input sentence of length $n$.
We can then simply run the Viterbi or the Forward algorithm on $G \cap S$ to compute Equartions~(\ref{eq:lognorm}) and~(\ref{eq:map}).
Note that $|E''| \propto n$, therefore the time-complexity is linear in the number of words.

We refer the reader to \citep{tapanainen1997intersection} for an introduction to this sequence tagging approach.

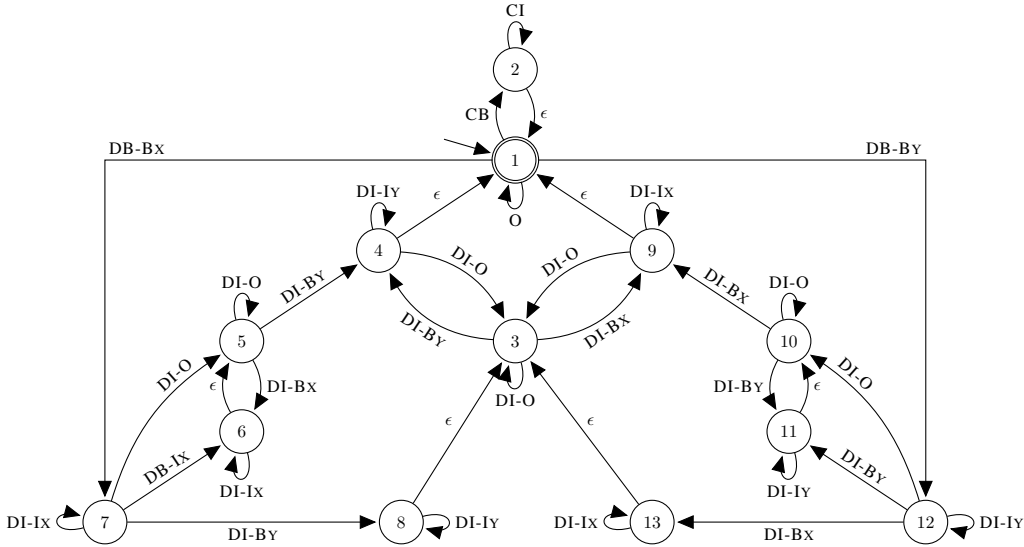
\begin{figure*}[t]
    \centering
    \input{figures/grammar}
    \caption{The grammar automaton we propose for discontinuous named-entity recognition.}
    \label{fig:fsa_disc}
\end{figure*}

\subsection{Grammar Automaton}

The grammar automaton used to constraint prediction to well-formed sequences of tags is shown in Figure~\ref{fig:fsa_disc}.
We present the automaton with $\epsilon$-transition for the sake of clarity, but they can be removed.
We omit weights as they are null.
States 1 and 2 recognize valid sequences of $\textsc{CB}$, $\textsc{CI}$ and $\textsc{O}$ tags.
Moreover, the structure of the WFSA states recognizing discontinuous mentions is symmetric: the left-hand (resp.\ right-hand) side recognizes discontinuous mentions whose leftmost component is typed $\textsc{x}$ (resp.\ $\textsc{y}$). Therefore we present only the left-hand side.

Transition $(1, \textsc{DB-Bx}, 7)$ starts the recognition of a set of mentions whose leftmost component is typed $\textsc{x}$.
The self-loop in state 7 recognizes following words of the first component.
Next we need to check that the inner structure of the set of mentions is well-formed.
On the one hand, states 5 and 6 allows to recognize following $\textsc{x}$ components and $\textsc{DI-O}$ tags,
until recognition of the first $\textsc{y}$ component via transition $(5, \textsc{DI-By}, 4)$.
On the other hand, transition $(7, \textsc{DB-Bx}, 8)$ starts the recognition of an component typed \textsc{y} that directly follows the first component.
Therefore, we need to check that there is ``something else'' in the set of mentions, otherwise the sequence of tags could lead to an ambiguity in the encoding of continuous mentions.
We ensure this via transition $(8, \epsilon, 3)$, that requires the generation of another component before reaching the final state.
Finally, states 3, 4 and 9 recognizes extra $\textsc{x}$ and $\textsc{y}$ in the set of mentions.

As such, the language of our grammar automaton is the set of well-formed tag sequences as described in Section~\ref{sec:scheme}.
To use our grammar automaton, we need to remove $\epsilon$-transitions.
The resulting WFSA has 22 states.\footnote{Altough 22 states is small and allows very fast computation, it is already too large for drawing a comprehensive figure.}
In the case of structural component types, we can simply remove transition $(1, \textsc{DB-By}, 12)$ to constrain the leftmost mention to be labeled \textsc{x}.

\textbf{Practical implementation.}
The intersection of the grammar and the sentence automata does not result in a homogeneous Markov chain as transition weights correspond to tag weights for the next word, and are therefore different at each step.
However, the resulting automaton has always a simple time-invariant structure.
In term of implementation, this reduces to applying a mask at each step, and both Viterbi and forward algorithms can be implemented using basic differentiable tensor operations.
For MAP inference, we compute the path of maximum weight and then rely on backpropagation to retrieve the sequence of tags \cite[][Section~2.1]{mensch2018diffdp}.

%% file: figures/grammar.tex
 \begin{tikzpicture}[
    shorten >=1pt,
    node distance=2cm,
    auto,
    > = triangle 45,
    % scaling
    scale=0.6,
    every node/.style={scale=0.6}
]

\node[state,accepting]  (q_1) at (0, 0) {$1$};
\node[state]  (q_2) at (0, 2) {$2$};
\node[state]  (q_3) at (0, -4) {$3$};

\node (start) at (-1.7, 0.5) {};
\path[->] (start) edge (q_1);

\path[->] (q_1) edge[bend left] node {\textsc{CB}} (q_2);
\path[->] (q_2) edge[bend left] node {$\epsilon$} (q_1);
\path[->] (q_2) edge[loop above] node {\textsc{CI}} (q_2);
\path[->] (q_1) edge[loop below] node {\textsc{O}} (q_1);
\path[->] (q_3) edge[loop below] node {\textsc{DI-O}} (q_3);

\node[state]  (q_4a) at (-3, -2) {$4$};
\node[state]  (q_5a) at (-6, -4) {$5$};
\node[state]  (q_6a) at (-6, -6) {$6$};
\node[state]  (q_7a) at (-9, -8) {$7$};
\node[state]  (q_8a) at (-2.5, -8) {$8$};

\draw[->] (q_1)  -| node[above,xshift=0.7cm] {\textsc{DB-Bx}} (q_7a);
\path[->] (q_7a) edge[bend left=20] node[sloped ,near end] {\textsc{DI-O}} (q_5a);
\path[->] (q_7a) edge[loop left] node {\textsc{DI-Ix}} (q_7a);
\path[->] (q_7a) edge node[sloped] {\textsc{DB-Ix}} (q_6a);
\path[->] (q_5a) edge[bend left] node {\textsc{DI-Bx}} (q_6a);
\path[->] (q_6a) edge[bend left] node {$\epsilon$} (q_5a);
\path[->] (q_5a) edge[loop above] node {\textsc{DI-O}} (q_5a);
\path[->] (q_6a) edge[loop below] node {\textsc{DI-Ix}} (q_6a);
\path[->] (q_5a) edge node[sloped] {\textsc{DI-By}} (q_4a);
\path[->] (q_4a) edge[loop above] node {\textsc{DI-Iy}} (q_4a);
\path[->] (q_4a) edge node {$\epsilon$} (q_1);
\path[->] (q_4a) edge[bend left] node[sloped] {\textsc{DI-O}} (q_3);
\path[->] (q_3) edge[bend left] node[sloped,below] {\textsc{DI-By}} (q_4a);
\path[->] (q_7a) edge node[below] {\textsc{DI-By}} (q_8a);
\path[->] (q_8a) edge node {$\epsilon$} (q_3);
\path[->] (q_8a) edge[loop right] node {\textsc{DI-Iy}} (q_8a);

\node[state]  (q_4b) at (3, -2) {$9$};
\node[state]  (q_5b) at (6, -4) {$10$};
\node[state]  (q_6b) at (6, -6) {$11$};
\node[state]  (q_7b) at (9, -8) {$12$};
\node[state]  (q_8b) at (3, -8) {$13$};

\draw[->] (q_1)  -| node[above,xshift=-0.7cm] {\textsc{DB-By}} (q_7b);
\path[->] (q_7b) edge[bend right=20,above] node[sloped,near end] {\textsc{DI-O}} (q_5b);
\path[->] (q_7b) edge node[sloped] {\textsc{DI-By}} (q_6b);
\path[->] (q_7b) edge[loop right] node {\textsc{DI-Iy}} (q_7b);
\path[->] (q_5b) edge[bend right,left] node {\textsc{DI-By}} (q_6b);
\path[->] (q_6b) edge[bend right,right] node {$\epsilon$} (q_5b);
\path[->] (q_5b) edge[loop above] node {\textsc{DI-O}} (q_5b);
\path[->] (q_6b) edge[loop below] node {\textsc{DI-Iy}} (q_6b);
\path[->] (q_5b) edge node[sloped] {\textsc{DI-Bx}} (q_4b);
\path[->] (q_4b) edge[loop above] node {\textsc{DI-Ix}} (q_4b);
\path[->] (q_4b) edge node[above] {$\epsilon$} (q_1);
\path[->] (q_4b) edge[bend right] node[sloped] {\textsc{DI-O}} (q_3);
\path[->] (q_3) edge[bend right] node[sloped,below] {\textsc{DI-Bx}} (q_4b);
\path[->] (q_7b) edge node[below] {\textsc{DI-Bx}} (q_8b);
\path[->] (q_8b) edge node[above,xshift=0.4em] {$\epsilon$} (q_3);
\path[->] (q_8b) edge[loop left] node {\textsc{DI-Ix}} (q_8b);

\end{tikzpicture}

%% file: 4.training.tex
\section{Weakly-Supervised Learning}
\label{sec:weaklysup}

The negative log-likelihood (NLL) loss,
\begin{align*}
\ell(\vw; \vy)
&= - \langle \vy, \vw \rangle + A_Y(\vw)\,,
\end{align*}
requires knowledge of the gold output $\vy$.
Unfortunately, NER datasets only contains annotated mentions, but not their component types
(\emph{e.g.}\ we do not know which components are body parts and events).
Therefore, we need to resort on weakly-supervised learning to infer this information.

\subsection{Learning with Partial Labels}

Learning with partial labels refers to the case where the gold output is unknown but there is access to a subset of labels that includes the gold one \cite{grandvalet2004partial,nguyen2008partial,cour2011partial}.
Let $\widetilde{Y} \subseteq Y$ be the set of tag sequences that recovers the gold discontinuous mentions.
For the example in Figure~\ref{fig:tag_sequence},
$\widetilde Y$ contain two sequences, one where components of the set of mentions are labeled $\textsc{x}/\textsc{x}/\textsc{y}/\textsc{y}$ and the other $\textsc{y}/\textsc{y}/\textsc{x}/\textsc{x}$.
For a sentence containing $k$ sets of mentions, we have $|\widetilde Y| = 2^k$.

Following \citet{jin2002partial}, we minimize the NLL after marginalizing over $\widetilde Y$:
\begin{align}
    &\widetilde{\ell}(\vw; \widetilde{Y})
    = -\log p_\theta(\widetilde{Y} | \vx)
    = -\log \sum_{\vy \in \widetilde{Y}} p_\theta(\vy | \vx) \nonumber
    \\
    &= A_Y(f_\theta(\vx)) -
        \underbrace{
            \log \sum_{\vy \in \widetilde{Y}} \exp\langle \vy, f_\theta(\vx) \rangle
            }_{
                = A_{\widetilde{Y}}(f_\theta(\vx))
            }\,,
    \label{eq:partial_loss}
\end{align}
where $A_{\widetilde Y}$ is the \emph{clamped} log-partition, which can be efficiently computed via a dynamic programming algorithm.
In speech processing, $A_{\widetilde Y}$ is called the alignment model and the associated FSA the numerator graph \cite{povey2016purely,hadian2018end}.

\textbf{Relation with EM.}
We can interpret minimizing $\widetilde\ell$ as an Expectation-Maximization (EM) procedure \cite{neal1998em}.
Indeed, the variational formulation of the clamped log-partition is:
\begin{align*}
    A_{\widetilde Y}(\vw) &= \sup_{\vmu \in \conv \widetilde{Y}} \langle \vmu, \vw\rangle - \Omega_{\widetilde{Y}}(\vmu)\,,
\end{align*}
where $\conv$ denotes the convex hull and $\Omega_{\widetilde{Y}}$ is a structured entropy term as described by \citet[Section~7.1]{blondel2020fy}.
Setting $\vw = f_\theta(\vx)$, by Danskin's theorem \cite{danskin1966,bertsekas1997nonlinear}, the gradient of the $A$ is:
\begin{align*}
    \widehat\vmu_{\widetilde Y}(\vw) = \nabla A_{\widetilde Y}(\vw) &= \argmax_{\vmu \in \conv \widetilde{Y}}~\langle \vmu, \vw\rangle - \Omega_{\widetilde{Y}}(\vmu)\,.
\end{align*}
We rewrite the minimization of $\widetilde\ell$ as a two-step procedure:
\begin{enumerate}
    \item \textbf{E step:} compute $\widehat\vmu_{\widetilde Y}(\vw)$;
    \item \textbf{M step:} take one gradient step over the network parameters using the marginal distribution computed in E step, yielding the loss:
    $$
        \ell(\vw; \widehat\vmu_{\widetilde Y}(\vw)) = - \langle \vy, \widehat\vmu(\vw) \rangle + A_Y(\vw)\,.
    $$
\end{enumerate}
It is important to note that $\widehat\vmu_{\widetilde Y}(\vw)$ is considered as a constant in the M step, \emph{i.e.}\ the gradient is:
\begin{align*}
    \nabla \ell(\vw; \widehat\vmu_{\widetilde Y}(\vw))\!=\!- \widehat\vmu(\vw)\!+\!\nabla A_Y(\vw)
    \!=\!\nabla \widetilde\ell(\vw;\widetilde Y)\,,
\end{align*}
meaning that this EM procedure is equivalent to minimizing the loss in Equation~(\ref{eq:partial_loss}).

This suggests a ``Hard EM'' alternative, where the $E$ step computes the unregularized maximum:
\begin{align*}
    \widehat\vy_{\widetilde Y}(\vw) = 
    \argmax_{\vy \in \conv \bar{Y}}~\langle \vy, \vw\rangle\,,
\end{align*}
and then apply one step of gradient descent using the loss $\ell(\vw; \widehat\vy_{\widetilde Y}(\vw))$ in the M step.

\subsection{Silver Annotation of Components}

In order to automatically annotate components,
we collect names of body parts from the metathesaurus \texttt{MRCONSO.RRF} of the Unified Medical Language System (UMLS, version 2023ab).\footnote{\url{https://www.ncbi.nlm.nih.gov/books/NBK9685/table/ch03.T.concept_names_and_sources_file_mr/}}
We select English entries corresponding to semantic types ``Body Location or Region'', ``Body Part, Organ, or Organ Component'' and ``Body Space or Junction'',
via the annotation in the lexicon \texttt{MRSTY.RRF},
which corresponds to identifiers 
\textsc{T029}, \textsc{T023} and \textsc{T030}, respectively.\footnote{\url{https://www.ncbi.nlm.nih.gov/books/NBK9685/table/ch03.Tf/}}
However, we remove all acronyms (indicated via the marker \textsc{Abr})
as they would introduce too many false positives in the annotation process (\emph{e.g.}\ ``in'' and ``am'' are acronyms of body parts).
This leads to $218\,134$ names of body parts.

Then, we try to match words of components with these entries.
If at least one word of a component match an entry, we consider it as a body part.
Note that a single match fully disambiguate a set of mentions.

%% file: 5.related_work.tex
\section{Related Work}
\label{sec:related_work}

\textbf{Tagging methods.}
\citet{tang2013dner_tagging} proposed the \textsc{BIOHD} tagging scheme for discontinuous NER.
A major issue of their approach is its \emph{structural ambiguity}: several tag sequences can encode the same discontinuous mention, and different discontinuous mentions have the same associated tag sequence, see \citep[Section~3.1]{muis2016graph}.
A choice to resolve ambiguity has to be made when making a prediction, meaning that there are structures that cannot be predicted.
Moreover, this approach does not constrain the output tag sequence to be well-formed, \emph{i.e.}\ it may not be possible to reconstruct mentions from a predicted tag sequence.
The tagging scheme used by \citet{metke2016concept} and \citet{dai2017tagging} has the same limitation.
\citet{muis2016graph} proposed a graph-based method that ensures that predictions are well-formed, but their approach still exhibits structural ambiguity.

\textbf{Other methods.}
\citet{wang2019combining} rely on a two-step model that first predicts continuous spans (\emph{i.e.}\ components) and then uses a separate classifier that combines them together.
\citet{dai2020dner_transition} proposed a novel transition-based model.
These two approaches are based on sequential predictions that are trained using gold intermediate outputs, which can lead to error propagation once a single mistake is made at test time.
To resolve this problem, \citet{wang2021dner_clique} proposed a method that jointly predicts spans and their combination based on the maximal clique problem.
A downside of these approaches is that they are more computationally costly (and therefore slower) than tagging methods.

%% file: 6.experiments.tex
\input{tables/results_f1}
\input{tables/data}

\input{tables/timing}

\section{Experiments}
\label{sec:exp}

We evaluate our approach on three standard English datasets for discontinuous named-entity recognition in the biomedical domain:
\cadec{} \cite{cadec}, \sha{} \cite{share2013} and \shb{} \cite{share2014}.
We pre-process the data using the script of \citet{dai2020dner_transition}.
Note that our tagging scheme cannot predict all discontinuous mentions in the data, \emph{i.e.}\ there are sentences that we cannot convert to our representation. Therefore, we remove these sentences from the training set.\footnote{Obviously, we do not remove anything from the test set.}
Data statistics are given in Table~\ref{tab:data_stats}.

\subsection{Discontinuity Analysis}

We conduct a qualitative analysis of the search space of our algorithm on the full \textsc{Cadec} dataset.
There are 26 discontinuous NER structures incompatible with our approach.\footnote{We do not count single mentions: we count full sets of mentions that cannot be recognized by our algorithm.}

There are discontinuous mentions where there is a \emph{partially} shared component.
This is due to shared negation (1 case), shared adjective (5 cases) and shared prepositional phrase (PP, 1 case):
\input{examples/incompatible1}
Although we cannot recognize these structures,
we could extend our automaton to recognize the shared part as a continuous chunk (negation, adjective or PP), and the rest using our two layer representation.

There are also discontinuous mentions that are composed of three components (16 cases), which we cannot recognize.
This can happens because there is a coordination in both subject and PP positions as in the following example:\footnote{This example has been slightly changed for formatting.}
\input{examples/incompatible2}
The mention ``\texttt{muscle aches in elbows}'' is composed of three components.

Finally, the last three incompatibilities are due to a convolated syntactic structure and annotation errors (2 cases).
Interestingly, some annotation errors can be detected thanks to our new annotation schema.
For example, in \cadec{} the sequence ``\texttt{renal and respiratory failure}'' as been incorrectly annotated as containing \texttt{renal respiratory failure} instead of \texttt{renal failure}.
In \shb{}, the sequence ``\texttt{pleural / abdominal effusions}'' as been incorrectly annotated as containing \texttt{effusions} instead of \texttt{abdominal effusions}.
Note that in this paper we used the datasets as such and did not fix any error so that results are comparable with previous work.

\subsection{Results}

Our neural network is excessively simple: we use the \textsc{Deberta-V3} pretrained self-attentive network \cite{he2021debertav3,he2021deberta} followed by a single linear projection that maps context-sensitive embeddings to tag weights.
All training details are given in Appendix~\ref{sec:training}.
For each loss function, we train six models with six different seeds and we select the best model using the development set.

\textbf{Results.}
We report the F-measure on all mentions and on discontinuous mentions only in Table~\ref{table:results}.
The evaluation is conducted on the the original representation so results are comparable with previous work.
Our approach leads to similar results to previous work.
We do not observe significant differences between different loss functions.

\textbf{Speed.}
All numbers are reported for computation on \textsc{Nvidia V100} GPUs.
Training takes approximately 40, 60 and 80 minutes on \cadec{}, \sha{} and \shb{}, respectively.
Table~\ref{tab:timing} compares decoding with previous work of \citet{dai2020dner_transition} and \citet{wang2021dner_clique}.
The transition-based model of \citet{dai2020dner_transition} is particularly slow as their approach cannot fully exploit GPU parallelization.
Our approach is $\sim$40-50 times faster that the method of \citet{wang2021dner_clique}.
This is due to two reasons: (1)~they use a complex neural network architecture on top of a BERT-like model and (2)~for each input they must solve a NP-hard problem (maximum clique) to make the prediction.

%% file: tables/results_f1.tex
\begin{table*}[t]
\small
\centering

\begin{tabular}{@{}l@{\hskip 0.3in}cc@{\hskip 0.3in}cc@{\hskip 0.3in}cc@{}}
\toprule
& \multicolumn{2}{c}{\cadec} & \multicolumn{2}{c}{\sha} & \multicolumn{2}{c}{\shb} \\
\cmidrule(l{0pt}r{2pt}){2-3}
\cmidrule(l{2pt}r{2pt}){4-5}
\cmidrule(l{2pt}r{0pt}){6-7}
& F1 & Disc.\ F1
& F1 & Disc.\ F1
& F1 & Disc.\ F1
\\
\midrule
\multicolumn{7}{l}{\textbf{Previous work}}
\\
\midrule
\citet{tang2013dner_tagging}
&&
&75.0& 
&&
\\

\citet{tang2018recognizing}
&66.3&
&&
&&
\\

\citet{metke2016concept}
& 64.4 &
& 56.5 &
& 60.2 &
\\
\citet{metke2016concept}$\dagger$
& 67.4 & 1.8
& 74.9  & 18.8 
& 76.6 & 6.0
\\

\citet{muis2016graph}$\dagger$
& 58.0 &  23.9
& 70.3 &  50.0
& 74.7 &  41.1
\\

\citet{dai2020dner_transition}
& 69.0  & 37.9
& 77.7 & 52.5
& 79.6 & 49.2
\\

\citet{wang2021dner_clique}
& 71.5 & 44.4
& 81.2 & 55.9
& 81.3 & 54.1
\\

\midrule
\multicolumn{7}{l}{\textbf{This work}} \\
\midrule
Soft EM & 71.1 & 38.1 & 80.7 & 49.2 & 81.5 & 51.9 \\
Hard EM & 71.9 & 35.9 & 82.0 & 51.9 & 81.6 & 54.1 \\
Weakly soft EM & 71.8 & 37.6 & 82.0 & 52.0 & 81.4 & 46.2 \\
Weakly hard EM & 70.4 & 33.6 & 82.0 & 52.1 & 81.8 & 49.8 \\
Structural labels & 72.9 & 41.5 & 82.1 & 53.3 & 80.9 & 53.7 \\
\bottomrule
\end{tabular}

\caption{Results on on three different datasets. Results marked with $\dagger$ are reproductions by \citet{wang2021dner_clique}.}
\label{table:results}
\end{table*}

%% file: tables/data.tex
\begin{table}[t]
\small
\centering
\begin{tabular}{lr@{}lr@{}lr@{}l}
    \toprule
    Split
    & \multicolumn{2}{l}{\cadec{}}
    & \multicolumn{2}{l}{\sha{}}
    & \multicolumn{2}{l}{\shb{}}
    \\
    \midrule
     Train
     & 5340 & ~(306)
     & 8508 & ~(477)
     & 17407 & ~(777)
     \\
     - filtered
     & 5322 & ~(288)
     & 8432 & ~(401)
     & 17294 & ~(667)
     \\
     Dev.
     & 1097 & ~(59)
     & 1250 & ~(58)
     & 1361 & ~(59)
     \\
     Test
     & 1160 & ~(74)
     & 9009 & ~(301)
     & 15850 & ~(411)
     \\
     \bottomrule
\end{tabular}
\caption{Number of sentences in each split. The number in parentheses corresponds to the number of sentences with at least one discontinuous mention.}
\label{tab:data_stats}
\end{table}

%% file: tables/timing.tex
\begin{table}[t]
    \small
    \centering
    \begin{tabular}{lrrr}
        \toprule
        Model
        & \cadec{}
        & \textsc{S2013} %\sharea{}
        & \textsc{S2014} %\shareb{}
        \\
        \midrule
         \citet{dai2020dner_transition}
         & 36
         & 41
         & 40
         \\
         \citet{wang2021dner_clique}
         & 193
         & 200
         & 198
         \\
         This work
         & 8286
         & 10216
         & 10206
         \\
         \bottomrule
    \end{tabular}
    \caption{
    Speed comparison in terms of sentence per seconds.
    Numbers for \citet{dai2020dner_transition} are \textsc{Bert}-based models, as reproduced by \citet{wang2021dner_clique}.
    }
    \label{tab:timing}
\end{table}

%% file: examples/incompatible1.tex
\begin{tikzpicture}[
    every node/.style={
        rectangle,
        inner xsep=0cm,
        inner ysep=0.1cm,
        text height=1.5ex,
        text depth=.25ex,
    }
]
    % weakness in arms and weakness in shoulders
    \node (couldnt) [rectangle] {\texttt{Couldn't}};
    \node (walk) [rectangle, right=0.2cm of couldnt] {\texttt{walk}};
    \node (or) [rectangle, right=0.2cm of walk] {\texttt{or}};
    \node (even) [rectangle, right=0.2cm of or] {\texttt{even}};
    \node (sleep) [rectangle, right=0.2cm of even] {\texttt{sleep}};
    \node (comfortably) [rectangle, right=0.2cm of sleep] {\texttt{comfortably}};

    \draw (couldnt.north west) --coordinate (part1)  (walk.north east);
    \draw (comfortably.north west) --coordinate (part2) (comfortably.north east);

    \coordinate[above=0.2cm of part1] (part1_b);
    \draw[dashed] (part1) -- (part1_b);
    \coordinate[above=0.2cm of part2] (part2_b);
    \draw[dashed] (part2) -- (part2_b);

    \draw[dashed] (part1_b) --node[yshift=+0.25cm] {\textsc{Adr}} (part2_b);

    \draw (couldnt.south west) --coordinate (part3)  (couldnt.south east);
    \draw (sleep.south west) --coordinate (part4) (comfortably.south east);

    \coordinate[below=0.2cm of part3] (part3_b);
    \draw[dashed] (part3) -- (part3_b);
    \coordinate[below=0.2cm of part4] (part4_b);
    \draw[dashed] (part4) -- (part4_b);

    \draw[dashed] (part3_b) --node[yshift=-0.25cm] {\textsc{Adr}} (part4_b);

\end{tikzpicture}
\begin{tikzpicture}[
    every node/.style={
        rectangle,
        inner xsep=0cm,
        inner ysep=0.1cm,
        text height=1.5ex,
        text depth=.25ex,
    }
]
    % weakness in arms and weakness in shoulders
    \node (severe) [rectangle] {\texttt{severe}};
    \node (colon) [rectangle, right=0.2cm of severe] {\texttt{colon}};
    \node (and) [rectangle, right=0.2cm of colon] {\texttt{and}};
    \node (uterine) [rectangle, right=0.2cm of and] {\texttt{uterine}};
    \node (cramping) [rectangle, right=0.2cm of uterine] {\texttt{cramping}};

    \draw (severe.north west) --coordinate (part1)  (colon.north east);
    \draw (cramping.north west) --coordinate (part2) (cramping.north east);

    \coordinate[above=0.2cm of part1] (part1_b);
    \draw[dashed] (part1) -- (part1_b);
    \coordinate[above=0.2cm of part2] (part2_b);
    \draw[dashed] (part2) -- (part2_b);

    \draw[dashed] (part1_b) --node[yshift=+0.25cm] {\textsc{Adr}} (part2_b);

    \draw (severe.south west) --coordinate (part3)  (severe.south east);
    \draw (uterine.south west) --coordinate (part4) (cramping.south east);

    \coordinate[below=0.2cm of part3] (part3_b);
    \draw[dashed] (part3) -- (part3_b);
    \coordinate[below=0.2cm of part4] (part4_b);
    \draw[dashed] (part4) -- (part4_b);

    \draw[dashed] (part3_b) --node[yshift=-0.25cm] {\textsc{Adr}} (part4_b);

\end{tikzpicture}
\null\hspace{-0.2cm}\begin{tikzpicture}[
    every node/.style={
        rectangle,
        inner xsep=0cm,
        inner ysep=0.1cm,
        text height=1.5ex,
        text depth=.25ex,
    }
]
    % weakness in arms and weakness in shoulders
    \node (muscle) [rectangle] {\texttt{muscle}};
    \node (fatigue) [rectangle, right=0.2cm of muscle] {\texttt{fatigue}};
    \node (and) [rectangle, right=0.2cm of fatigue] {\texttt{/}};
    \node (soreness) [rectangle, right=0.2cm of and] {\texttt{soreness}};
    \node (in) [rectangle, right=0.2cm of soreness] {\texttt{in}};
    \node (my) [rectangle, right=0.2cm of in] {\texttt{my}};
    \node (forearms) [rectangle, right=0.2cm of my] {\texttt{forearms}};

    \draw (muscle.north west) --coordinate (part1)  (fatigue.north east);
    \draw (in.north west) --coordinate (part2) (forearms.north east);

    \coordinate[above=0.2cm of part1] (part1_b);
    \draw[dashed] (part1) -- (part1_b);
    \coordinate[above=0.2cm of part2] (part2_b);
    \draw[dashed] (part2) -- (part2_b);

    \draw[dashed] (part1_b) --node[yshift=+0.25cm] {\textsc{Adr}} (part2_b);

    \draw (muscle.south west) --coordinate (part3)  (muscle.south east);
    \draw (soreness.south west) --coordinate (part4) (forearms.south east);

    \coordinate[below=0.2cm of part3] (part3_b);
    \draw[dashed] (part3) -- (part3_b);
    \coordinate[below=0.2cm of part4] (part4_b);
    \draw[dashed] (part4) -- (part4_b);

    \draw[dashed] (part3_b) --node[yshift=-0.25cm] {\textsc{Adr}} (part4_b);
\end{tikzpicture}

%% file: examples/incompatible2.tex
\null\hspace{-0.4cm}\begin{tikzpicture}[
    every node/.style={
        rectangle,
        inner xsep=0cm,
        inner ysep=0.1cm,
        text height=1.5ex,
        text depth=.25ex,
    }
]
    \node (muscle) [rectangle] {\texttt{muscle}};
    \node (and) [rectangle, right=0.2cm of muscle] {\texttt{and}};
    \node (joint) [rectangle, right=0.2cm of and] {\texttt{joint}};
    \node (aches) [rectangle, right=0.2cm of joint] {\texttt{aches}};
    \node (in) [rectangle, right=0.2cm of aches] {\texttt{in}};
    \node (arms) [rectangle, right=0.2cm of in] {\texttt{arms}};
    \node (and2) [rectangle, right=0.2cm of arms] {\texttt{and}};
    \node (elbows) [rectangle, right=0.2cm of and2] {\texttt{elbows}};

    \draw(joint.north west)--node[yshift=+0.25cm]{\textsc{Adr}} (arms.north east);
    
% 90 Muscle aches in arms     (0, 0, 3, 5)

    \draw ($(muscle.north west)+(0,16pt)$) --coordinate (part1)  ($(muscle.north east)+(0,16pt)$);
    \draw ($(aches.north west)+(0,16pt)$) --coordinate (part2) ($(arms.north east)+(0,16pt)$);

    \coordinate[above=0.2cm of part1] (part1_b);
    \draw[dashed] (part1) -- (part1_b);
    \coordinate[above=0.2cm of part2] (part2_b);
    \draw[dashed] (part2) -- (part2_b);

    \draw[dashed] (part1_b) --node[yshift=+0.25cm] {\textsc{Adr}} (part2_b);

% 91 joint aches in elbows    (2, 4, 7, 7)

    \draw ($(joint.north west)+(0,36pt)$) --coordinate (part3)  ($(in.north east)+(0,36pt)$);
    \draw ($(elbows.north west)+(0,36pt)$) --coordinate (part4) ($(elbows.north east)+(0,36pt)$);

    \coordinate[above=0.2cm of part3] (part3_b);
    \draw[dashed] (part3) -- (part3_b);
    \coordinate[above=0.2cm of part4] (part4_b);
    \draw[dashed] (part4) -- (part4_b);

    \draw[dashed] (part3_b) --node[yshift=+0.25cm] {\textsc{Adr}} (part4_b);

% 93 Muscle aches in elbows   (0, 0, 3, 4, 7, 7)

    \draw (muscle.south west) --coordinate (part5)  (muscle.south east);
    \draw (aches.south west) --coordinate (part6) (in.south east);
    \draw (elbows.south west) --coordinate (part7) (elbows.south east);

    \coordinate[below=0.2cm of part5] (part5_b);
    \draw[dashed] (part5) -- (part5_b);
    \coordinate[below=0.2cm of part6] (part6_b);
    \draw[dashed] (part6) -- (part6_b);
    \coordinate[below=0.2cm of part7] (part7_b);
    \draw[dashed] (part7) -- (part7_b);

    \draw[dashed] (part5_b) --node[yshift=-0.25cm] {\textsc{Adr}} (part7_b);

\end{tikzpicture}

%% file: 7.conclusion.tex
\section{Conclusion}

In this work, we propose a novel tagging scheme for discontinuous NER based on a two-layer representation of discontinuous mentions.
Our approach leads to result on par with state-of-the-art using a very simple neural network architecture.
Importantly, decoding with our model is very fast compared to previous work.

Our main objective with this work is to propose a simple plug-in method for discontinuous NER: any future work on models for BIO tagging can now also be trivially evaluated on discontinuous NER.
Moreover, our approach is also fast to train, meaning that there is no significant cost overhead.

%% file: 10.nn.tex
\section{Training details}
\label{sec:training}

The model is trained for 20 epochs using the cosine learning rate scheduler as implemented in the HuggingFace library.
The maximum learning rate is fixed to $10^{-5}$.
The warmup ratio is $10\%$.
We apply dropout with a probability of $0.5$ to BERT's output.
The gradient norm is clipped to 1.
All parameters have a weight decay of $0.01$.
We use the Adam variant proposed by \citet{mosbach2021adamw}.